\pdfoutput=1
\documentclass{article}
\usepackage{spconf,amsmath,graphicx}
\usepackage{amssymb}
\usepackage{hyperref}
\usepackage[skip=2pt]{caption}

\title{CO-OCCURRENCE MATRIX ANALYSIS-BASED SEMI-SUPERVISED TRAINING FOR OBJECT DETECTION}
\name{ \parbox{\linewidth}{\centering
Min-Kook Choi$^1$, Jaehyeong Park$^1$, Jihun Jung$^1$, Heechul Jung$^2$, Jin-Hee Lee$^1$, \\ Woong Jae Won$^1$, Woo Young Jung$^1$, Jincheol Kim$^3$, and Soon Kwon$^{1*}$\thanks{$^*$Corresponding author.}}}
\address{DGIST, Daegu, Republic of Korea$^1$ \\ KAIST, Daejeon, Republic of Korea$^2$ \\ SK Telecom, Seoul, Republic of Korea$^3$}
\begin{document}
%
\maketitle
\begin{abstract}
One of the most important factors in training object recognition networks using convolutional neural networks (CNNs) is the provision of annotated data accompanying human judgment. Particularly, in object detection or semantic segmentation, the annotation process requires considerable human effort. In this paper, we propose a semi-supervised learning (SSL)-based training methodology for object detection, which makes use of automatic labeling of un-annotated data by applying a network previously trained from an annotated dataset. Because an inferred label by the trained network is dependent on the learned parameters, it is often meaningless for re-training the network. To transfer a valuable inferred label to the unlabeled data, we propose a re-alignment method based on co-occurrence matrix analysis that takes into account one-hot-vector encoding of the estimated label and the correlation between the objects in the image. We used an MS-COCO detection dataset to verify the performance of the proposed SSL method and deformable neural networks (D-ConvNets) \cite{jifeng17} as an object detector for basic training. The performance of the existing state-of-the-art detectors (D-ConvNets, YOLO v2 \cite{redmon17}, and single shot multi-box detector (SSD) \cite{liu16}) can be improved by the proposed SSL method without using the additional model parameter or modifying the network architecture. 
\end{abstract}

\section{Introduction}
\label{sec:intro}

\begin{figure}[ht!]
\begin{center}
\includegraphics[width=8.5cm]{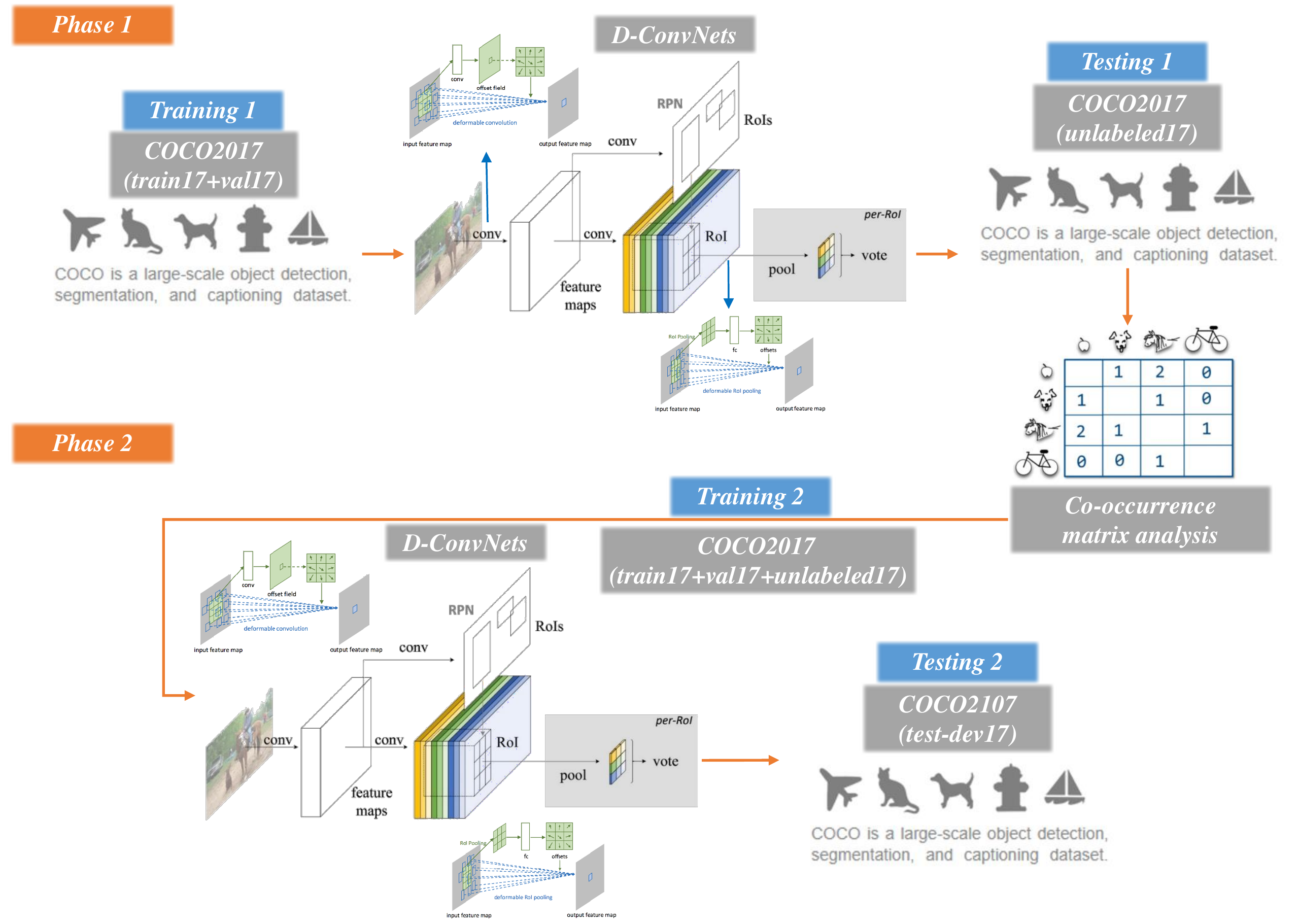}\\
\caption{\textbf{Overview of the proposed SSL pipeline.} The proposed technique consists of two steps. In the first step, the detector for labeling the unlabeled data is trained with the existing annotated data (Training1), and then the inferring process for the unlabeled data is performed (Testing1). After performing pseudo-labeling through one-hot-vector encoding and co-occurrence matrix analysis, a new network is trained (Training2) and the data for evaluation are inferred (Testing2).}
\label{fig:overview}
\end{center}
\end{figure}

Recently, the effectiveness of convolutional neural networks (CNNs) in the object detection has been improved, and the performance of the object detectors that has stagnated since the appearance of the Histogram of Oriented Gradient (HOG) based the detector \cite{felzenszwalb09} has been greatly advanced. There are two main types of end-to-end training object detectors that utilize CNNs as a backbone architecture \cite{he16, huang17, sabour17}. There are two-stage networks of region-based detectors with a Network in Network (NIN) structure by training the region candidates from the region proposal network \cite{jifeng17, jifeng16, ren15}, and one-stage networks that learn the region of the objects from sub-regions in predefined areas \cite{redmon17, liu16, lin17}. Both types of networks have played a significant role in improving the dramatic performance of CNN and the decoder network for multi-tasking.

Despite the dramatic improvement in performance of state-of-the-art detectors, object detectors trained by machine learning techniques have the disadvantage of having a large capacity for the refined datasets for training. Currently, the annotation method widely used for the production of learning data provides a simple user interface that can specify the class of the object to be classified and the position of the bounding box, and it can be used in a crowdsourcing platform \cite{olga15, lin14}. Notwithstanding the evolution of these platforms and methodologies, annotation techniques that rely on human labor are still a burden on learning algorithms. Particularly, annotation cost is a big obstacle to learn a good network for object detection and semantic segmentation \cite{lin14, cordts16}.

Various efforts have been made to overcome these problems in object classification and detection. Lee \cite{lee13} proposed a simple pseudo-labeling technique to utilize the learned network for semi-supervised learning (SSL). Although the idea of pseudo-labeling of trained networks has long been proposed, re-training with pseudo-labeling depends on the parameters of the learned network, making it difficult to obtain improved results in re-training. In order to solve this problem, Lee proposed a weight control-based learning method for the pseudo labeled data in cross-entropy loss and confirmed the possibility of SSL technique using the learned network. Yan et al. \cite{yan17} proposed an object detector using EM (Expectation-Maximization (EM)) to apply the SSL-based training method to the object detection. They proposed an algorithm that updates the CNN internal parameters from the probabilities of the inferred data for non-label data whereas the object detector learns through the EM algorithm.

In this paper, we propose a simple but powerful one-hot-vector encoding based on the SSL idea and a semi-supervised training method through co-occurrence matrix analysis. The latest performance networks deduce a bounding box of the correct form that can be used as training data in a specific object or visual environment. However, if we use the result of inference as a pseudo label in direct way, we cannot obtain the big learning effect by dependency of parameter and data. In order to compensate for the effect of pseudo-labeling during training, 1) the inference result is encoded as a one-hot-vector and 2) the co-occurrence matrix obtained from the prior knowledge is used to recalculate whether the inference result is suitable for training. Through these two steps, it is decided whether the inferred bounding box is included in the training dataset and the new network is learned through the updated dataset. Figure \ref{fig:overview} outlines the proposed SSL learning scheme. As a result of testing the SSL scheme with the MS-COCO detection dataset, we confirmed the performance improvement in the state-of-the-art detectors such as deformable neural networks (D-ConvNets) \cite{jifeng17}, YOLO v2 \cite{redmon17}, and single shot multi-box detector (SSD) \cite{liu16} in terms of accuracy using mean average precision (mAP) without any additional parameter or architecture modification.

\section{Semi-supervised learning with deformable neural networks}
\label{sec:ssl}

\subsection{Deformable convolutional networks}
\label{ssec:dconvnets}

To apply our SSL method, we utilize the D-ConvNets object detector \cite{jifeng17}, which uses the CNN combining with deformable operation and achieves state-of-the-art performance with the MS-COCO detection evaluation dataset \cite{lin14}. For kernel weight of a general CNN, learnable convolutional parameters are only learned for neighboring pixels or its atrous spatial position \cite{chen16} at every pixel location. In this case, there is a limitation that the kernel weight at one pixel location is only considered to be neighboring with local neighbors. D-ConvNets does not limit the pixel location of the kernel weights to be learned by adding the deformable offset parameter to the learnable pixel location.

The learning objectives of D-ConvNets are defined as follows:
\begin{equation}
\label{eq:dconv}
y(p_0) = \sum_{p_n \in R} w(p_n)\cdot x(p_0+p_n+\bigtriangleup p_n); 
\end{equation}
where $w$ is the kernel weight in the network, $x$ is the input of the network at a particular layer, $R$ is a regualr grid over the input, and $p_0$ is the 2D coordinate position of the kernel weights to be learned. $\bigtriangleup p_n$ is a newly introduced learnable offset parameter through which a deformable element of a convolution-capable region is introduced to help learn various types of kernels that appear in the detection of objects and objects with severe deformation. In order to improve the performance of object detection, deformable operation is applied to a few top layers of the backbone CNN of the region-based fully convolutional networks (R-FCN) \cite{jifeng16} and the position sensitive ROI pooling layer for localization.

\subsection{Pseudo labeling with one-hot-vector encoding}
\label{ssec:ssl}

We use a one-hot-vector based pseudo-labeling technique to train D-ConvNets, which is learned as the baseline algorithm for the proposed SSL using unlabeled images. In order to perform pseudo-labeling, the inferred bounding boxes with the softmax output higher than the threshold value of the inferred bounding box of D-ConvNets are encoded as one-hot vector to provide learnable pseudo-label that will be used for later training.

\begin{equation}
\label{eq:one-hot-vector}
LB(i) =
  \begin{cases}
    [\hat{x},\hat{y},\hat{w},\hat{h},c]       & \quad \text{if } \frac{\exp{q_j}}{\sum_{j=1}^{n}\exp(q_j)} > \rho \\
    [\, ]  & \quad \text{otherwise}, 
  \end{cases}
\end{equation}
$LB(\cdot)$ is the index dictionary for the training label and has input the vector in the form $[x, y, w, h, c]$, and $i$ is an $i$th new labeled object in an inferred image to add previously annotated dataset. In this case, $x, y, w, h$ represent the position and size of the bounding box and $c$ is a class label corresponding to the softmax output. $q_j$ is the inferred responses of last layer in D-ConvNets and $n$ is the total number of desired classes to detect with given dataset. If it exceeds the given threshold value $\rho$, the pseudo-label obtained through the assigned label is used for learning together with the previously annotated data in the future. 

\subsection{Co-occurrence matrix analysis}
\label{ssec:co_occur}

In order to maximize the efficiency of our SSL, we propose the use of a co-occurrence matrix that is extracted by prior knowledge of annotated data. Co-occurrence matrix is a matrix of the probability that objects in the image appear on the same image. Because inferring the probability of existence of a specific object with only the learned object detector is biased with regard to the training result, it is effective to represent the relation with co-occurred objects to readjust the probability of inference and use it for pseudo-labeling.

\begin{figure}[t!]
\begin{center}
\includegraphics[width=8.5cm]{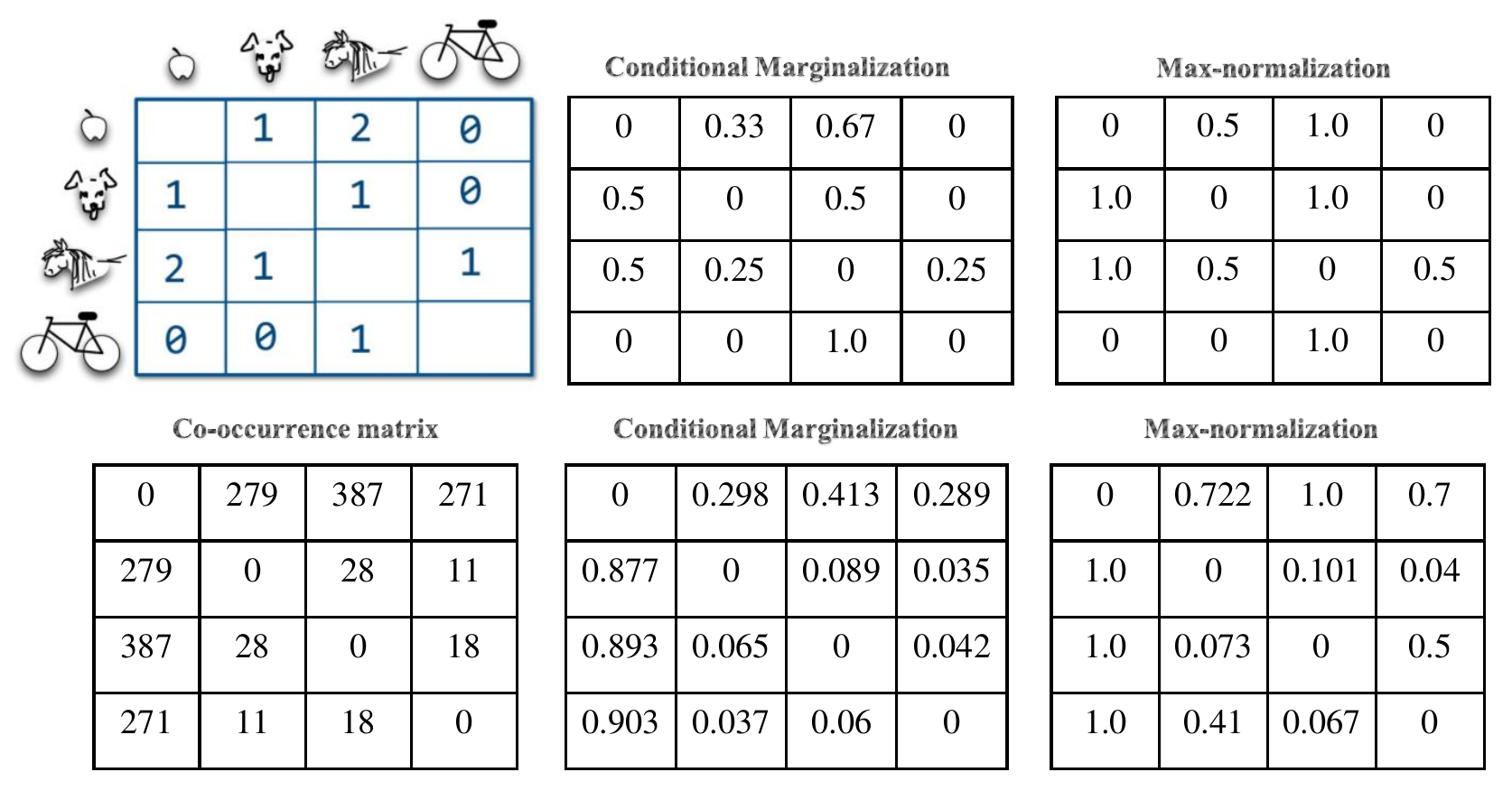}\\
\caption{\textbf{Example of co-occurrence matrix.} A co-occurrence matrix of four classes with five images (top) and a case with prior knowledge from a large-scale dataset (bottom) using the conditional marginalization and the max-normalization.}
\label{fig:co-occurrence}
\end{center}
\end{figure}

To represent the relationship between objects in a form that can be calculated when constructing the co-occurrence matrix, conditional independence between objects is assumed. Then, to normalize the strong relationship between the objects in the image, max-normalization was performed. Finally, in order to efficiently apply the information on the prior knowledge after the maximum normalization, only the relation between the two strongest objects is applied to the final softmax output correction when several objects exist in the image at the same time.

\begin{equation}
\label{eq:prob}
\begin{aligned}
p(x|z_1,z_2,\cdots ,z_n) = \prod^{n}_{i=1}p(x|z_1, \cdots ,z_n)  \\
\approx \max{p(x|z_i)}, \forall i=1,2, \cdots,n,
\end{aligned}
\end{equation}
where $n$ is the total number of classes to detect. For example, if there are four classes (see Figure \ref{fig:co-occurrence}) of desired objects to detect in an image, we can apply a rule in Equation \ref{eq:prob} to extract co-occurrence probability of the class apple as following: $p(apple|dog,horse,bike) \approx p(apple|horse)$. To reflect the extracted correction probabilities, we need to re-scale the inferred softmax probability for pseudo-labeling with the co-occurrence matrix values. In this case, the labeling to unlabeled data to which the co-occurrence threshold is applied in Equation \ref{eq:one-hot-vector} is defined as follows. 

\begin{equation}
\label{eq:co-occur}
LB(i) =
  \begin{cases}
    [\hat{x},\hat{y},\hat{w},\hat{h},c]       & \quad \text{if } \frac{\exp{q_j}}{\sum_{j=1}^{n}\exp(q_j)} \cdot \sigma > \rho_{co} \\
    [\, ]  & \quad \text{otherwise}, 
  \end{cases}
\end{equation}
where $\sigma=\max{p(x|z_i)}$ is the probability for desired class object from co-occurrence matrix and $\rho_{co}$ is the threshold for pseudo-labeling the same as Equation \ref{eq:one-hot-vector} which applies one-hot-vector encoding with an inferred output. 

\section{Experimental results}
\label{sec:exp}

We used the MS-COCO detection dataset \cite{lin14} to verify the effectiveness of the proposed SSL method. The MS-COCO dataset provides training and testing data and evaluation tools for visual recognition applications such as object detection and instance segmentation, key-point detection, scene parsing, and unlabeled2017 data for unsupervised or semi-supervised learning. 

\begin{figure}[t!]
\begin{center}
\includegraphics[width=8.7cm]{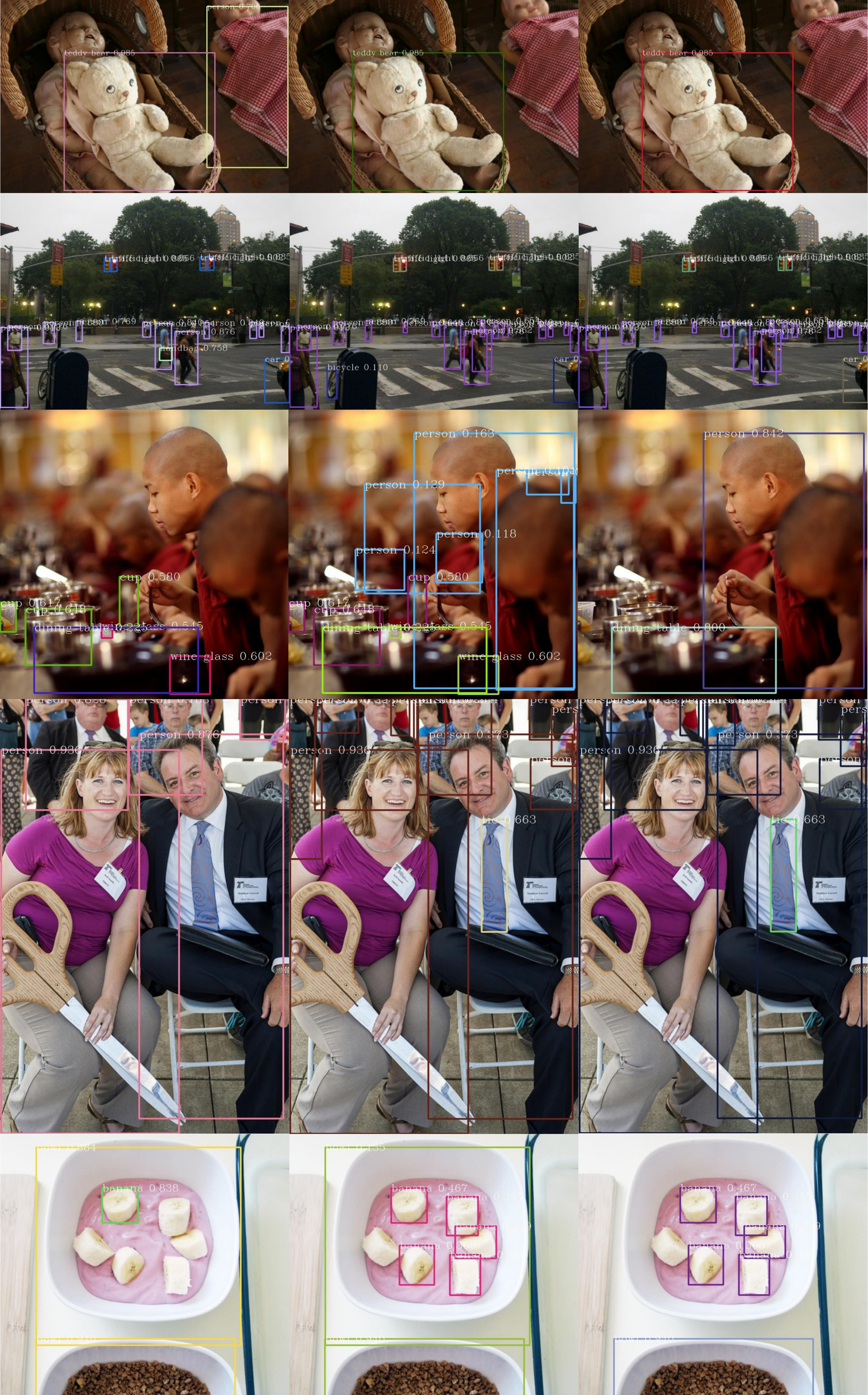}\\
\caption{\textbf{Examples of pseudo-labeling results.} From the left, the results of the basic model ($\rho=0.5$), applying co-occurrence matrix with ($\rho=0.5$, $\rho_{co}=0.1$), and ($\rho=0.5$, $\rho_{co}=0.3$). There is a large difference in the result of pseudo-labeling according to the set threshold value. For the first row, we could remove the bounding box for the mis-inferred object, and for the second and third rows, we detected additional objects in the complex scene. The fourth row detected a small tie object, which is difficult to deduce in a complex scene, based on the relation between objects. The final row detected additional bounding boxes of undetected objects. }
\label{fig:pseudo}
\end{center}
\end{figure}

In order to verify the proposed SSL method, we used a single model of D-ConvNets as a baseline detection architecture for initial pseudo-labeling. Among the recently proposed CNN architectures \cite{he16, huang17, sabour17}, pre-trained ResNet-101 \cite{he16} with ImageNet was used as the backbone CNN for the baseline architecture training. In the baseline model training, input data size are $1200 \times 800$, and the total training epoch sets are 10 and the learning rate starts from $5 \times 10^{-3}$. We apply $10^{-1}$ times dropping scale in 5.3 epoch. To apply the proposed learning metric, we need to set the threshold parameter $\rho$ for initial pseudo-labeling and another threshold parameter $\rho_{co}$ for the co-occurrence matrix analysis. The $\rho$ for one-hot-vector encoding in Training1 was set to $0.5$ and $0.7$, and the $\rho$ required for Training2 was set to 0.5, and for $\rho_{co}$, $0.1$, $0.2$, $0.3$, and $0.4$ were set (see Figure \ref{fig:overview}). Figure \ref{fig:pseudo} shows the inference results of pseudo-labeling for the proposed SSL from a single model of trained D-ConvNets. When the co-occurrence matrix was readjusted, the accuracy of pseudo-labeling could be corrected according to the set threshold value. When the set threshold value is high, only the conservative reasoning result is included in the training dataset.

Table \ref{tb:param} shows quantitative results obtained by applying a number of learned model to the test-dev17 data according to the set threshold value. For evaluation of the MS-COCO detection dataset, the mean absolute precision (mAP) is obtained by increasing the intersection on union (IoU) of the inferred bounding box versus the ground truth bounding box by $0.05$ from $0.5$ to $0.95$ ($[0.5: 0.05: 0.95]$). According to Table \ref{tb:param}, the highest mAP was recorded at $\rho=0.5$ and $\rho_{co}=0.3$. 

Table \ref{tb:model} shows the results of applying the proposed SSL method to different detectors \cite{jifeng17, redmon17, liu16} with the SSL parameters obtained from Table \ref{tb:param}. In order to train SSD, the backbone CNN used ResNet-101 with ImageNet, which is already trained, and the training detail is as follows. SGD was used for training optimization, and the learning rate was started at $10^{-3}$,and dropping scale $10^{-1}$ was applied at 80 k, 100 k, and 120 k. The total learning epoch is 32, and the scale minimum ratio for the default box is set to 10. The input size of the image was re-scaled to 512$\times$512, the momentum was set to 0.9, and the weight decay was set to $5 \times 10^{-4}$. The backbone CNN for training YOLO v2 utilizes the previously trained Darknet-19 with ImageNet, and the training details are as follows. For the optimization, SGD is used same as SSD, and the learning rate starts from $10^{-3}$ and the dropping scale $10^{-1}$ is applied at 266 and 300 epoch. The total learning epoch is 500, and the training metric like \cite{redmon17} is applied for high-resolution images with 608$\times$608. As a result, using the proposed SSL technique, we can confirm that the performance of the YOLO v2 and SSD is improved over 1.0 mAP in [0.5: 0.05: 0.95] in our evaluations.

\begin{table}[t!]
\centering
\caption{MS-COCO detection dataset evaluations for [0.5:0.05:0.95] using D-ConvNets with different parameters.}
\label{tb:param}
\begin{tabular}{c|c}
\hline
Model (backbone, SSL parameter(s), \\ training dataset) & mAP \\
\hline
\hline
D-ConvNets (ResNet-101, none, \\ train17 $+$ val17) & 36.3 \\
\hline
D-ConvNets (ResNet-101, $\rho=0.5$, \\ train17 $+$ val17 $+$ unlabeled17) & 37.0 \\
\hline
D-ConvNets (ResNet-101, $\rho=0.7$, \\ train17 $+$ val17 $+$ unlabeled17) & 36.7 \\
\hline
D-ConvNets (ResNet-101, $\rho=0.5$, $\rho_{co}=0.1$, \\ train17 $+$ val17 $+$ unlabeled17) & 37.6 \\
\hline
D-ConvNets (ResNet-101, $\rho=0.5$, $\rho_{co}=0.2$, \\ train17 $+$ val17 $+$ unlabeled17) & 37.3 \\
\hline
{\bf D-ConvNets} (ResNet-101, $\rho=0.5$, $\rho_{co}=0.3$, \\ train17 $+$ val17 $+$ unlabeled17) & {\bf 37.8} \\
\hline
D-ConvNets (ResNet-101, $\rho=0.5$, $\rho_{co}=0.4$, \\ train17 $+$ val17 $+$ unlabeled17) & 37.5 \\
\hline
\end{tabular}
\end{table}

\begin{table}[t!]
\centering
\caption{MS-COCO detection dataset evaluations for [0.5:0.05:0.95] using different architectures with or without the proposed SSL ($\rho=0.5$, $\rho_{co}=0.3$).}
\label{tb:model}
\begin{tabular}{c|c|c}
\hline
Model (backbone CNN) & mAP & mAP with SSL\\
\hline
\hline
SSD \cite{liu16} (ResNet-101) & 24.1 & \bf{25.3} \\
\hline
YOLO v2 \cite{redmon17} (Darknet-19) & 24.0 & \bf{25.1} \\
\hline
D-ConvNets \cite{jifeng17} (ResNet-101) & 36.3 & \bf{37.8}\\
\hline
\end{tabular}
\end{table}

\smallskip

\section{Conclusion}
\label{sec:con}

In this paper, we proposed a method to improve object detection using SSL. The proposed SSL scheme has the advantage of automatically acquiring trainable labeled data without any additional human effort to insert a new annotation. We also proposed a metric to improve the performance of existing one-hot-vector-based SSL using a co-occurrence matrix. When training is performed applying the proposed SSL technique, the learning time is increased in accordance with the increased amount of data, but the improved performance can be expected without modification of the architecture or additional parameters. 

In order to further clarify the performance of our SSL scheme, it is necessary to verify various models according to the setting of the hyper-parameter of the single model, and the influence of the SSL scheme on the network type such as one stage or two stages and the difference between them. In addition, there is a need to analyze what effect the baseline detector has on performance.


\bibliographystyle{IEEEbib}
\bibliography{icip2018_final}

\begin{thebibliography}{10}

\bibitem{jifeng17}
J.~Dai, H.~Qi, Y.~Xiong, Y.~Li, G.~Zhang, H.~Hu, and Y.~Wei,
\newblock ``Deformable convolutional networks,''
\newblock in {\em International Conference on Computer Vision (ICCV)}, 2017.

\bibitem{redmon17}
J.~Redmon and A.~Farhadi,
\newblock ``Yolo9000: Better, faster, stronger,''
\newblock in {\em Computer Vision and Pattern Recognition (CVPR)}, 2017.

\bibitem{liu16}
W.~Liu, D.~Anguelov, D.~Erhan, C.~Szegedy, S.~Reed, C.~Y. Fu, and A.~C. Berg,
\newblock ``Ssd: Single shot multibox detector,''
\newblock in {\em European Conference on Computer Vision (ECCV)}, 2016.

\bibitem{felzenszwalb09}
P.~F. Felzenszwalb, R.~B. Girshick, D.~McAllester, and D.~Ramanan,
\newblock ``Object detection with discriminatively trained part-based models,''
\newblock {\em IEEE Transactions on Pattern Analysis and Machine Intelligence},
  vol. 32, no. 9, pp. 1627--1645, 2009.

\bibitem{he16}
K.~He, X.~Zhang, S.~Ren, and J.~Sun,
\newblock ``Deep residual learning for image recognition,''
\newblock in {\em Computer Vision and Pattern Recognition (CVPR)}, 2016.

\bibitem{huang17}
G.~Huang, Z.~Liu, L.~van~der Maaten, and K.~Q. Weinberger,
\newblock ``Densely connected convolutional networks,''
\newblock in {\em Computer Vision and Pattern Recognition (CVPR)}, 2017.

\bibitem{sabour17}
S.~Sabour, N.~Frosst, and G.~E. Hinton,
\newblock ``Dynamic routing between capsules,''
\newblock in {\em Advances in Neural Information Processing Systems (NIPS)},
  2017.

\bibitem{jifeng16}
J.~Dai, Y.~Li, K.~He, and J.~Sun,
\newblock ``R-fcn: Object detection via region-based fully convolutional
  networks,''
\newblock in {\em Advances in Neural Information Processing Systems (NIPS)},
  2016.

\bibitem{ren15}
S.~Ren, K.~He, R.~Girshick, and J.~Sun,
\newblock ``Faster r-cnn: Towards real-time object detection with region
  proposal networks,''
\newblock in {\em Advances in Neural Information Processing Systems (NIPS)},
  2015.

\bibitem{lin17}
T.~Y. Lin, P.~Goyal, R.~Girshick, K.~He, and P.~Dollár,
\newblock ``Focal loss for dense object detection,''
\newblock in {\em International Conference on Computer Vision (ICCV)}, 2017.

\bibitem{olga15}
O.~Russakovsky, J.~Deng, H.~Su, J.~Krause, S.~Satheesh, S.~Ma, Z.~Huang,
  A.~Karpathy, A.~Khosla, M.~Bernstein, A.~C. Berg, and L.~Fei-Fei,
\newblock ``Imagenet large scale visual recognition challenge,''
\newblock {\em International Journal of Computer Vision (IJCV)}, vol. 115, no.
  3, pp. 211--252, 2015.

\bibitem{lin14}
T.~Y. Lin, M.~Maire, S.~Belongie, L.~Bourdev, R.~Girshick, J.~Hays, P.~Perona,
  D.~Ramanan, C.~L. Zitnick, and P.~Dollár,
\newblock ``Microsoft coco: Common objects in context,''
\newblock in {\em European Conference on Computer Vision (ECCV)}, 2014.

\bibitem{cordts16}
M.~Cordts, M.~Omran, S.~Ramos, T.~Rehfeld, M.~Enzweiler, R.~Benenson,
  U.~Franke, S.~Roth, and B.~Schiele,
\newblock ``The cityscapes dataset for semantic urban scene understanding,''
\newblock in {\em Computer Vision and Pattern Recognition (CVPR)}, 2016.

\bibitem{lee13}
D.~Lee,
\newblock ``Pseudo-label: The simple and efficient semi-supervised learning
  method for deep neural networks,''
\newblock in {\em International Conference on Machine Learning Workshop
  (ICML-WS)}, 2013.

\bibitem{yan17}
Z.~Yan, J.Liang, W.~Pan, J.~Li, and C.~Zhang,
\newblock ``Weakly- and semi-supervised object detection with
  expectation-maximization algorithm,''
\newblock {\em CoRR}, vol. abs/1702.08740, 2017.

\bibitem{chen16}
L.~C. Chen, G.~Papandreou, I.~Kokkinos, K.~Murphy, and A.~L. Yuille,
\newblock ``Deeplab: Semantic image segmentation with deep convolutional nets,
  atrous convolution, and fully connected crfs,''
\newblock {\em CoRR}, vol. abs/1606.00915, 2016.

\end{thebibliography}

\end{document}